\documentclass[11pt,a4paper]{article}
\usepackage[hyperref]{emnlp2018}
\usepackage{times}
\usepackage{latexsym}
\usepackage{basecommon}

\usepackage{url}
\usepackage[]{caption}

\aclfinalcopy

\newcommand\fornn{\overset{\rightarrow}}
\newcommand\barnn{\overset{\leftarrow}}

\title{Entity Tracking Improves Cloze-style Reading Comprehension}

\author{Luong Hoang \quad \quad \quad Sam Wiseman  \quad \quad \quad Alexander M. Rush \\
         School of Engineering and Applied Sciences \\ Harvard University \\ Cambridge, MA, USA \\ 
         {\tt lhoanger@gmail.com, \{swiseman,srush\}@seas.harvard.edu } \\}

\date{}

\begin{document}
\maketitle
\begin{abstract}

Reading comprehension tasks test the ability of models to process long-term context
and remember salient information. Recent work has shown that relatively simple 
neural methods such as the Attention Sum-Reader can perform well on these tasks; 
however, these systems still significantly trail human performance. 
Analysis suggests that many of the remaining hard instances are related
to the inability to track entity-references throughout documents. 
This work focuses on these hard entity tracking cases with two extensions: (1) additional entity features, and (2) training with a multi-task tracking objective.
We show that these simple modifications improve performance both independently and in combination, and we outperform the previous state of the art on the LAMBADA dataset, particularly on difficult entity examples. 
\end{abstract}

\section{Introduction}
\label{sec:intro}
There has been tremendous interest over the past several years in Cloze-style~\citep{taylor1953cloze} reading comprehension tasks, datasets, and models~\citep{hermann2015teaching,hill16goldilocks,kadlec2016text,dhingra2016gated,cui2016attention}. 
Many of these systems apply neural models to learn to predict answers based on contextual matching, and have inspired other work in long-form generation and question answering. The extent and limits of these successes have also been a topic of interest~\citep{chen2016a,chu17broad}. Recent analysis by \citet{chu17broad} suggests that a significant portion of the errors made by standard models, especially on the LAMBADA dataset~\citep{paperno16the}, derive from the inability to correctly track entities or speakers, or a failure to handle various forms of reference. 

This work targets these shortcomings by designing a model and training scheme targeted towards \textit{entity tracking}. 
Specifically we introduce two simple changes to a stripped down model: (1) 
 simple, entity-focused features, and (2) two multi-task objectives that target entity tracking. 
Our ablation analysis shows that both independently improve entity tracking, which is the primary source of overall model's improvement.  Together they lead to state-of-the-art performance on LAMBADA dataset and near state-of-the-art on CBT dataset ~\citep{hill16goldilocks}, even with a relatively simple model. 

\setlength{\fboxsep}{1pt}

\begin{figure}[t]
 \small
\centering
\fbox{\includegraphics[scale=0.28]{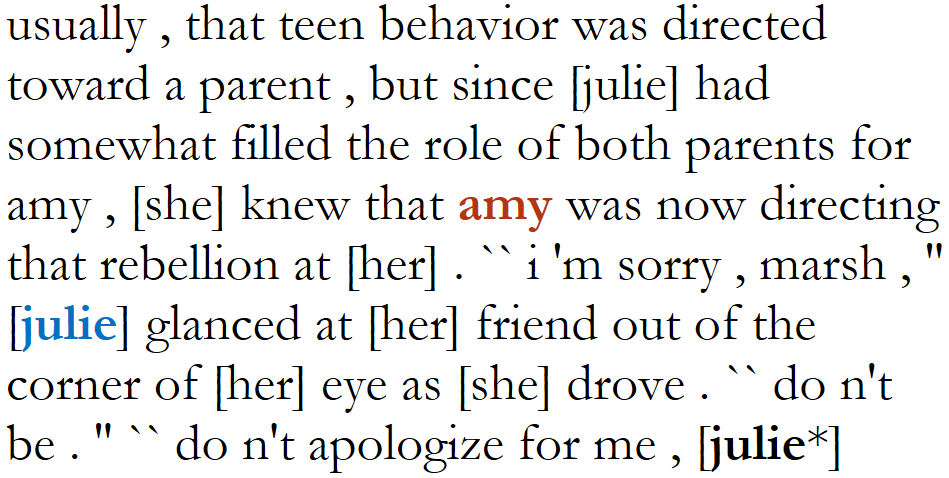}}
\caption{A LAMBADA example where the final word ``julie'' (with reference chain in brackets) is the answer, $y$, to be predicted from the preceding context $x$. 
A system must know the two speakers and the current dialogue turn, simple context matching is not sufficient. Here, our model's predictions \textcolor{red}{before} and \textcolor{blue}{after} adding multi-task objective are shown. 
}
\label{fig:rcexamp}
\end{figure}


\section{Background and Related Work}
Cloze-style reading comprehension uses a passage of word tokens $x = x_{1:n}$ (the \textit{context}), with one token $x_j$ masked; the task is to fill in the masked word $y$, which was originally at position $j$. These datasets aim to present a benchmark challenge requiring some understanding of the context to select the correct word. This task is a prerequisite for problems like long-form generation and document-based question answering. 

A number of datasets in this style exist with different focus. Here we considered the LAMBADA dataset and the named entity portion of the Children's Book Test dataset (CBT-NE). LAMBADA uses novels where examples consist of 4-5 sentences and the last word to be predicted is masked, $x_n$. The dataset is constructed carefully to focus on examples where humans needed the context to predict the masked word.
CBT-NE examples, on the other hand, include 21 sentences where the masked word is a named entity extracted from the last sentence, with $j \leq n$, and is constructed in a more automated way. We show an example from LAMBADA in Figure~\ref{fig:rcexamp}. In CBT, as well as the similar CNN/Daily Mail dataset~\cite{hermann2015teaching}, the answer $y$ is always contained in $x$ whereas in LAMBADA it may not be. \citet{chu17broad} showed, however, that training only on examples where $y$ \textit{is} in $x$ leads to improved overall performance, and we adopt this approach as well.





\paragraph{Related Work}
The first popular neural network reading comprehension models were the Attentive Reader and its variant Impatient Reader \citep{hermann2015teaching}. Both were the first to use bidirectional LSTMs to encode the context paragraph and the query separately. The Stanford Reader \citep{chen2016a} is a simpler version with fewer layers for inference. These models use an encoder to map each context token $x_i$ to a vector $\boldu_i$. 
Following the terminology of \citet{wang2017emergent}, \textit{explicit reference models} calculate a similarity measure $s_i = s(\boldu_i, \boldq)$ between each context vector $\boldu_i$ and a query vector $\boldq$ derived for the masked word. These similarity scores are projected to an attention distribution $\balpha = \mathrm{softmax}(\{s_i\})$ over the context positions in $1, \ldots, n$, which are taken to be candidate answers.

The Attention Sum Reader \citep{kadlec2016text} is a further simplified version. It computes $\boldu_i$ and $\boldq$ with separate bidirectional GRU~\cite{chung2014empirical} networks, and $s_i$ with a dot-product. 
It is trained to minimize:

\vspace{-0.5cm}
\begin{align*} 
\mcL^0(\btheta) &= -\ln p(y \given x, \boldq) \nonumber \\
&= -\ln \sum_{i:x_i = y} p(x_i \given \boldq) = -\ln \sum_{i:x_i = y} \alpha_i,
\end{align*}

\noindent where $\btheta$ is the set of all parameters associated with the model, and $y$ is the correct answer. At test time, a \textit{pointer sum attention} mechanism is used to predict the word type with the highest aggregate attention as the answer. The Gated Attention Reader \citep{dhingra2016gated} leverages the same mechanism for prediction and introduces an attention gate to modulate the joint context-query information over multiple hops. 

The Recurrent Entity Networks \citep{henaff2016tracking} uses a custom gated recurrent module, Dynamic Memory, to learn and update entity representations as new examples are received. Their gate function is combined of (1) a similarity measure between the input and the hidden states, and (2) a set of trainable "key" vectors which could learn any attribute of an entity such as its location or other entities it is interacting with in the current context. The Query Reduction Networks \citep{seo2016query} is also a gated recurrent network which tracks state in a paragraph and uses a hidden query vector to keep pointing to the answer at each step. The query is successively transformed with each new sentence to a reduced state that's easier to answer given the new information. 

\paragraph{Model}
In this work, we were particularly interested in the shortcomings of simple models and exploring whether or how much entity tracking could help, since \citet{chu17broad} has pointed out this weakness. As a result, we adapt a simplified Attention Sum (AttSum) reader throughout all experiments. Our version uses only a single bidirectional GRU for both $\boldu_i$ and $\boldq$. This GRU is of size $2d$, using the first $d$ states for the context and second $d$ for the query. Formally, let $\fornn{\boldh}_i$ and $\barnn{\boldh}_i$ (both in $\reals^{2d}$) represent the forward and backward states of a bidirectional GRU run over $x$, and let $\fornn{\boldh}_{i, \uparrow}$ and $\fornn{\boldh}_{i, \downarrow}$ be the first and second $d$ 
states respectively, and define $\vert \vert$ as the concatenation operator. The context vectors are constructed as $\boldu_i = \fornn{\boldh}_{i,\uparrow} \vert \vert \barnn{\boldh}_{i,\uparrow}$. For datasets using the last word, the query is constructed as $\boldq = \fornn{\boldh}_{n,\downarrow} \vert \vert \barnn{\boldh}_{1,\downarrow}$. When the masked word can be anywhere, the query is constructed as $\boldq = \fornn{\boldh}_{j-1,\downarrow} \vert \vert \barnn{\boldh}_{j+1,\downarrow}$.

Our main contribution is the extension of this simple model to incorporate entity tracking. Other authors have explored extending neural reading comprehension models with linguistic features, particularly \citet{dhingra2017linguistic} who use a modified GRU with knowledge such as coreference relations and hypernymy. In \citet{dhingra2018neural}, the most recent coreferent antecedent for each token is incorporated into the update equations of the GRU unit to bias the reader towards coreferent recency. In this work, we instead use a much simpler set of features and compare to this and several other models as baseline approaches.

\section{Learning to Track Entities}

Analysis on reading comprehension has indicated that neural models are strong at matching local context information but weaker at following entities through the discourse~\citep{chen2016a,chu17broad}.  We consider two straightforward ways for extending the Attention Sum baseline to better track entities.

\begin{table}[t]
\small
\centering
\begin{tabular}{cl}
\toprule
1 & Sentence Index, POS Tag, NER Tag \\
2 & Is among last 3 PERSON words in $\boldx$ \\  
3 & Is a PERSON word in the last sentence \\
4 & Is a PERSON word identical to previous PERSON word \\
5 & Is a PERSON word identical to next PERSON word \\
6 & Quoted-speech Index \\
7 & Speaker \\
\bottomrule
\end{tabular}
\caption{Word-level features used in AttSum-Feat model.}
\label{tab:features}
\end{table}

\paragraph{Method 1: Features}
We introduce a short-list of features in Table~\ref{tab:features} to augment the representation of each word in $x$. These features are meant to help the system to identify and use the relationships between words in the passage.\footnote{POS tags are produced with the NLTK library~ \citep{bird2009natural}, and NER tags with the Stanford NER tagger~\citep{finkel2005incorporating}. We additionally found it useful to tag animate words as PERSONs on the CBT-NE data, using the animate word list of \citet{bergsma2006bootstrapping}.} Features 2-5 apply only to words tagged PERSON by the NER tagger. Features 6-7 apply only to words between opening and closing quotation marks. Feature 6 indicates the index of the quote in the document, and Feature 7 gives the assumed speaker of the quote using some simple rules; we provide the rules in the Supplementary Material. Though most of these features are novel, they are motivated by recent analysis~\citep{wang2015machine,chen2016a,wang2017emergent}.

All features are incorporated into a word's representation by embedding each discrete feature into a vector of the same size as the original word embedding, adding the vectors as well as a bias, and applying a $\tanh$ nonlinearity.


\paragraph{Method 2: Multitasking }


We additionally encourage the neural 
model to keep track of entities by multitasking with simple auxiliary entity-tracking 
tasks. Examples such as Figure~\ref{fig:rcexamp} suggest that keeping track of which entities are currently in scope is useful for answering reading comprehension questions. There, \textit{amy} and \textit{julie} are conversing, and being able to track that \textit{amy} is the speaker of the final quote helps to rule her out as a candidate answer. We consider two tasks:






For Task 1 ($\mcL^1$) we train the same model to predict repeated named entities. For all named entities $x_j$ such that there is a $x_i = x_j$ with $i < j$, we attempt to mask and predict the word type $x_j$. This is done by introducing another Cloze prediction, but now setting the target $y=x_j$, reducing the context to preceding words $x_{1:j-1}$ with $\mathbf{u_i} = \fornn{\boldh}_{i}$ , and 
the query $\boldq = \fornn{\boldh}_{j-1}$.  (Note that unlike above, 
both of these only use the forward states of the GRU). 
We use a bilinear similarity score $s_i = \boldq^{\trans} \, \boldQ \, \boldu_i$, for this prediction 
where $\boldQ$ is a learned transformation in $\reals^{2d \times 2d}$. This task is inspired by the antecedent ranking task in coreference~\citep{wiseman15learning,wiseman16learning}.

For Task 2 ($\mcL^2$) we train to predict the order index in 
which a named entity has been introduced. For example, in Figure~~\ref{fig:rcexamp},  \textit{julie} would be 1, \textit{amy} would be 2, \textit{marsh} would be 3, etc.  
The hope here is that learning to predict when  entities reappear will help the model track their reoccurences. For the blue labeled \textit{julie}, the model would aim to 
predict 1, even though it appears later in the context.
This task is inspired by the One-Hot Pointer Reader of \citet{wang2017emergent} on the Who-did-What dataset \citep{onishi2016did}.
Formally, letting $\widehat{\mathrm{idx}}(x_j)$ be the predicted index for $x_j$, we minimize:

\vspace*{-0.5cm}
\begin{align*} 
\mcL^2(\btheta) &= -\ln p(\widehat{\mathrm{idx}}(x_j) \, {=} \, \mathrm{idx}(x_j) \given x_{1:j-1}) \nonumber \\
&= -\ln \mathrm{softmax}(\boldW \fornn{\boldh}_j)_{\mathrm{idx}(x_j)},
\end{align*}

\noindent where $\boldW \in \reals^{|\mcE| \times 2d}$ and $\mcE$ is the set of entity word types in the document. 
Note that this is a simpler computation, requiring  only $O(|\mcE| \times n)$ predictions per $\boldx$, whereas $\mcL^1$ requires $O(n^2)$.

The full model minimizes a multi-task loss: $\mcL^0(\btheta) + \gamma_1 \mcL^1(\btheta) + \gamma_2 \mcL^2(\btheta)$. Using $\mcL^1$ and $\mcL^2$ simultaneously did not lead to improved performance however, and so either $\gamma_1, \gamma_2$ is always 0. We believe that this is because, while the learning objectives for $\mcL^1$ and $\mcL^2$ are mathematically different, they are both designed to similarly track the entities mentioned so far in the document and thus do not provide complementary information to each other. 

We found it useful to have two hyperparameters per auxiliary task governing the number of
distinct named entity word \textit{types} and \textit{tokens} used in defining the losses $\mcL^1$ and $\mcL^2$. In particular, per document these hyperparameters control in a top-to-bottom order the number of distinct named entity word types we attempt to predict, as well as the number of tokens of each type considered. 


\section{Experiments}
\label{sec:experiments}

\paragraph{Methods}
\label{sec:methods}
This section highlights several aspects of our methodology; full hyperparameters are given in the Supplementary Material. For the training sets, we exclude examples where the answer is not in the context. The validation and test sets are not modified however and the model with the highest accuracy on the validation set is chosen for testing. For both tasks, the context words are mapped to learned embeddings; importantly, we initialize the first 100 dimensions with the 100-dimensional GLOVE embeddings ~\cite{pennington2014glove}. 
Named entity words are anonymized, as is done in the CNN/Daily Mail corpus~\cite{hermann2015teaching} and in some of the experiments of \citet{wang2017emergent}. The model is regularized with dropout~\citep{srivastava2014dropout} and optimized with  ADAM~\citep{kingma2014adam}. For all experiments we performed a random search over hyperparameter values~\citep{bergstra2012random}, and report the results of the models that performed best on the validation set. Our implementation is available at \url{https://github.com/harvardnlp/readcomp}.

\begin{table}[t!]
\small
\centering
\begin{tabular}{lcc}
\toprule
LAMBADA & Val & Test \\
\midrule
 
GA Reader \citep{chu17broad} & - & 49.00 \\
MAGE (48) \citep{dhingra2017linguistic} & 51.10 & 51.60 \\
MAGE (64) \citep{dhingra2017linguistic} & 52.10 & 51.10 \\
GA + C-GRU \citep{dhingra2018neural} & - & 55.69 \\
\midrule
AttSum & 56.03 & 55.60 \\
AttSum + $\mcL^1$ & 58.35 & 56.86 \\
AttSum + $\mcL^2$ & 58.08 & 57.29 \\
AttSum-Feat & 59.62 & 59.05 \\
AttSum-Feat + $\mcL^1$  & \textbf{60.22} & \textbf{59.23} \\
AttSum-Feat + $\mcL^2$  & 60.13 & 58.47 \\
\toprule
CBT-NE \\
\midrule 
GA Reader \citep{dhingra2016gated} & 78.50 & \textbf{74.90} \\
EpiReader \citep{trischler2016natural} & 75.30 & 69.70 \\
DIM Reader \citep{Liu2017} & 77.10 & 72.20 \\
AoA \citep{cui2016attention} & 77.80 & 72.0 \\
AoA + Reranker \citep{cui2016attention} & \textbf{79.60} & 74.0 \\
\midrule
AttSum & 74.35 & 69.96 \\
AttSum + $\mcL^1$ & 76.20 & 72.16 \\
AttSum + $\mcL^2$ & 76.80 & 72.60 \\
AttSum-Feat & 77.80 & 72.36 \\
AttSum-Feat + $\mcL^1$  & 78.40 & 74.36 \\
AttSum-Feat + $\mcL^2$  & 79.40 & 72.40 \\
\bottomrule
\end{tabular}
\caption{Validation \& Test results on all datasets. AttSum* are our models, 
including variants with features and multi-task loss. Others indicate previous best published results. All improvements over AttSum are statistically significant ($\alpha$ = 0.05) according to the McNemar test with continuity correction \citep{dietterich1998approximate}.}
\label{tab:mainresults}
\end{table}

\paragraph{Results and Discussion}



Table~\ref{tab:mainresults} shows the full results of our best models on the LAMBADA and CBT-NE datasets, and compares them to recent, best-performing results in the literature. 

For both tasks the inclusion of either entity features or multi-task objectives leads to large statistically significant increases in validation and test score, according to the McNemar test ($\alpha = 0.05$) with continuity correction \citep{dietterich1998approximate}. Without features, AttSum + $\mcL^2$ achieves the best test results, whereas with features  AttSum-Feat + $\mcL^1$ performs best on CBT-NE. The results on LAMBADA indicate that entity tracking is a very important overlooked aspect of the task. Interestingly, with features included, AttSum-Feat + $\mcL^2$ appears to hurt test performance on LAMBADA and leaves CBT-NE performance essentially unchanged, amounting to a negative result for $\mcL^2$. On the other hand, the effect of AttSum-Feat + $\mcL^1$ is pronounced on CBT-NE, and while our simple models do not increase the state-of-the-art test performance on CBT-NE, they outperform ``attention-over-attention'' in addition to reranking~\cite{cui2016attention}, and is outperformed only by architectures supporting ``multiple-hop'' inference over the document~\cite{dhingra2016gated}. Our best model on CBT-NE test set, AttSum-Feat + $\mcL^1$, is very close to the current state-of-the-art result. On the validation sets for both LAMBADA and CBT-NE, the improvements from adding features to AttSum + $\mcL^i$ are statistically significant (for full results refer to our supplementary material). On LAMBADA, the $\mcL^1$ multi-tasked model is a
3.5-point increase on the state of the art.



Our method also employs fewer parameters than other richer models such as the GA Reader in \citep{dhingra2016gated}. More specifically, in terms of number of parameters, our models are very similar to a 1-hop GA Reader. In contrast, all published experiments of the latter use 3 hops where each hop requires 2 separate Bi-GRUs, one to model the document and one for the query. This constitutes the largest difference in model size between the two approaches.

Table \ref{tab:lamb_ablat} considers the performance of the different models based on a segmentation of the data. Here we consider examples where: (1) Entity - if the answer is a named entity; (2) Speaker - if the answer is a named entity and the speaker of quote; (3) Quote - if the answer is found within a quoted speech. Note that Speaker and Quote categories, while mutually exclusive, are subsets of the overall Entity category. We see that both the additional features and multi-task objectives independently result in a clear improvement in all categories, but that the gains are particularly pronounced for named entities and specifically for Speaker and Quote examples. Here we see sizable increases in performance, particularly in the Speaker category. We see larger increases in the more dialog heavy LAMBADA task.

\begin{figure}[t]
 \small
\centering
\fbox{\includegraphics[scale=0.26]{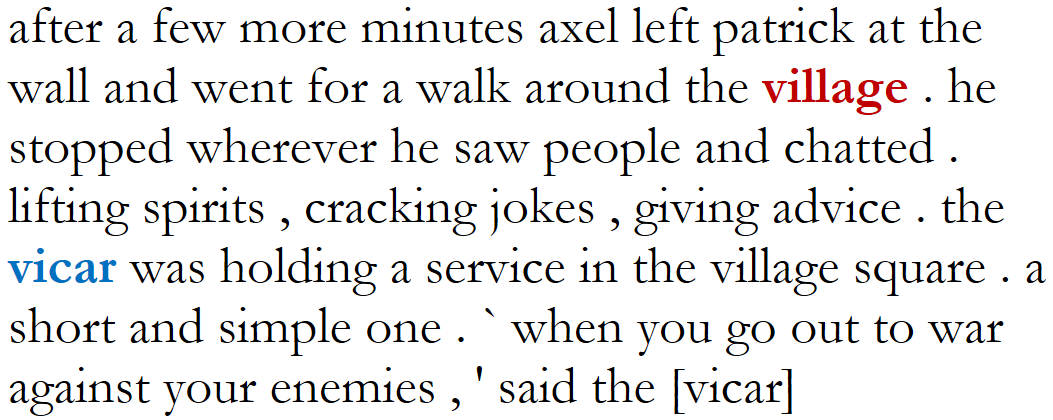}}
\fbox{\includegraphics[scale=0.26]{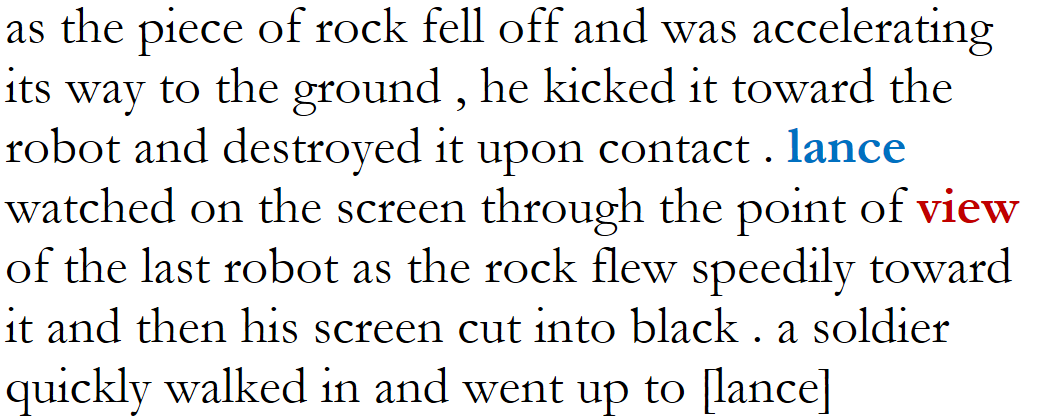}}
\caption{LAMBADA examples where AttSum incorrectly predicts a non-entity answer whereas AttSum-Feat and AttSum + $\mcL^i$ choose correctly.  
}
\label{fig:nonentexamp}
\end{figure}

As a qualitative example of the improvement afforded by multi-task training, in Figure \ref{fig:rcexamp} we show the different predictions made by our model with and without $\mcL^1$ (colored as blue and red, respectively). Note that \textit{amy} and \textit{julie} are both entities that have been repeated twice in the passage. In addition to the final answer, our model with the $\mcL^1$ loss was also able to predict these entities (at the colored locations) given preceding words. Further qualitative analysis reveals that these augmentations improved the model's ability to eliminate non-entity choices from predictions. Some examples are shown in Figure~\ref{fig:nonentexamp}.

\begin{table}[t!]
\small
\centering
\begin{tabular}{lcccccc}
\toprule
LAMBADA & All & Entity & Speaker &  Quote \\
\midrule
AttSum                 & 56.03 & 75.17 & 74.81 & 73.31\\
AttSum      + $\mcL^1$ & 58.35 & 78.51 & 78.38 & 79.42\\
AttSum      + $\mcL^2$ & 58.08 & 78.17 & 77.96 & 76.76\\
AttSum-Feat            & 59.62 & 79.40 & 80.34 & 79.68\\
AttSum-Feat + $\mcL^1$ & 60.22 & 82.00 & 82.98 & 81.67\\
AttSum-Feat + $\mcL^2$ & 60.14 & 82.06 & 83.06 & 82.60\\
\midrule
CBT-NE \\
\midrule
AttSum                 & 74.35 & 76.28 & 75.08 & 74.96\\
AttSum      + $\mcL^1$ & 76.20 & 78.03 & 76.98 & 77.33\\
AttSum      + $\mcL^2$ & 76.80 & 77.45 & 76.27 & 76.48\\
AttSum-Feat            & 77.80 & 80.58 & 79.84 & 79.61\\
AttSum-Feat + $\mcL^1$ & 78.40 & 80.44 & 79.68 & 79.78\\
AttSum-Feat + $\mcL^2$ & 79.40 & 82.41 & 81.51 & 81.39\\
\bottomrule
\end{tabular}
\caption{Ablation results on validation sets, see text for definitions of the numeric columns and models.}
\label{tab:lamb_ablat}
\end{table}




\section{Conclusion}
This work demonstrates that learning to track entities with features and multi-task learning significantly increases the performance of a baseline reading comprehension system, particularly on the difficult LAMBADA dataset. This result indicates that higher-level word relationships may not be modeled by simple neural systems, but can be incorporated with minor additional extensions. This work hints that it is difficult for vanilla models to learn long-distance entity relations, and that these may need to be encoded directly through features or possibly with better pre-trained representations.

\section*{Acknowledgments}
\begin{small}
SW gratefully acknowledges the support of a Siebel Scholars award. AMR gratefully acknowledges the support of NSF CCF-1704834, Intel Research, and Amazon AWS Research grants.
\end{small}

\bibliography{emnlp2018}

\begin{thebibliography}{29}
\expandafter\ifx\csname natexlab\endcsname\relax\def\natexlab#1{#1}\fi

\bibitem[{Bergsma and Lin(2006)}]{bergsma2006bootstrapping}
Shane Bergsma and Dekang Lin. 2006.
\newblock Bootstrapping path-based pronoun resolution.
\newblock In \emph{Proceedings of the 21st International Conference on
  Computational Linguistics and the 44th annual meeting of the Association for
  Computational Linguistics}, pages 33--40. Association for Computational
  Linguistics.

\bibitem[{Bergstra and Bengio(2012)}]{bergstra2012random}
James Bergstra and Yoshua Bengio. 2012.
\newblock Random search for hyper-parameter optimization.
\newblock \emph{Journal of Machine Learning Research}, 13(Feb):281--305.

\bibitem[{Bird et~al.(2009)Bird, Klein, and Loper}]{bird2009natural}
Steven Bird, Ewan Klein, and Edward Loper. 2009.
\newblock \emph{Natural language processing with Python: analyzing text with
  the natural language toolkit}.
\newblock " O'Reilly Media, Inc.".

\bibitem[{Chen et~al.(2016)Chen, Bolton, and Manning}]{chen2016a}
Danqi Chen, Jason Bolton, and Christopher~D. Manning. 2016.
\newblock A thorough examination of the cnn/daily mail reading comprehension
  task.
\newblock In \emph{{ACL}}.

\bibitem[{Chu et~al.(2017)Chu, Wang, Gimpel, and McAllester}]{chu17broad}
Zewei Chu, Hai Wang, Kevin Gimpel, and David~A. McAllester. 2017.
\newblock Broad context language modeling as reading comprehension.
\newblock In \emph{{EACL}}, pages 52--57.

\bibitem[{Chung et~al.(2014)Chung, Gulcehre, Cho, and
  Bengio}]{chung2014empirical}
Junyoung Chung, Caglar Gulcehre, KyungHyun Cho, and Yoshua Bengio. 2014.
\newblock Empirical evaluation of gated recurrent neural networks on sequence
  modeling.
\newblock \emph{arXiv preprint arXiv:1412.3555}.

\bibitem[{Cui et~al.(2016)Cui, Chen, Wei, Wang, Liu, and Hu}]{cui2016attention}
Yiming Cui, Zhipeng Chen, Si~Wei, Shijin Wang, Ting Liu, and Guoping Hu. 2016.
\newblock Attention-over-attention neural networks for reading comprehension.
\newblock \emph{arXiv preprint arXiv:1607.04423}.

\bibitem[{Dhingra et~al.(2018)Dhingra, Jin, Yang, Cohen, and
  Salakhutdinov}]{dhingra2018neural}
Bhuwan Dhingra, Qiao Jin, Zhilin Yang, William~W Cohen, and Ruslan
  Salakhutdinov. 2018.
\newblock Neural models for reasoning over multiple mentions using coreference.
\newblock \emph{arXiv preprint arXiv:1804.05922}.

\bibitem[{Dhingra et~al.(2016)Dhingra, Liu, Yang, Cohen, and
  Salakhutdinov}]{dhingra2016gated}
Bhuwan Dhingra, Hanxiao Liu, Zhilin Yang, William~W Cohen, and Ruslan
  Salakhutdinov. 2016.
\newblock Gated-attention readers for text comprehension.
\newblock \emph{arXiv preprint arXiv:1606.01549}.

\bibitem[{Dhingra et~al.(2017)Dhingra, Yang, Cohen, and
  Salakhutdinov}]{dhingra2017linguistic}
Bhuwan Dhingra, Zhilin Yang, William~W Cohen, and Ruslan Salakhutdinov. 2017.
\newblock Linguistic knowledge as memory for recurrent neural networks.
\newblock \emph{arXiv preprint arXiv:1703.02620}.

\bibitem[{Dietterich(1998)}]{dietterich1998approximate}
Thomas~G Dietterich. 1998.
\newblock Approximate statistical tests for comparing supervised classification
  learning algorithms.
\newblock \emph{Neural computation}, 10(7):1895--1923.

\bibitem[{Finkel et~al.(2005)Finkel, Grenager, and
  Manning}]{finkel2005incorporating}
Jenny~Rose Finkel, Trond Grenager, and Christopher Manning. 2005.
\newblock Incorporating non-local information into information extraction
  systems by gibbs sampling.
\newblock In \emph{Proceedings of the 43rd annual meeting on association for
  computational linguistics}, pages 363--370. Association for Computational
  Linguistics.

\bibitem[{Henaff et~al.(2016)Henaff, Weston, Szlam, Bordes, and
  LeCun}]{henaff2016tracking}
Mikael Henaff, Jason Weston, Arthur Szlam, Antoine Bordes, and Yann LeCun.
  2016.
\newblock Tracking the world state with recurrent entity networks.
\newblock \emph{arXiv preprint arXiv:1612.03969}.

\bibitem[{Hermann et~al.(2015)Hermann, Kocisky, Grefenstette, Espeholt, Kay,
  Suleyman, and Blunsom}]{hermann2015teaching}
Karl~Moritz Hermann, Tomas Kocisky, Edward Grefenstette, Lasse Espeholt, Will
  Kay, Mustafa Suleyman, and Phil Blunsom. 2015.
\newblock Teaching machines to read and comprehend.
\newblock In \emph{Advances in Neural Information Processing Systems}, pages
  1693--1701.

\bibitem[{Hill et~al.(2016)Hill, Bordes, Chopra, and Weston}]{hill16goldilocks}
Felix Hill, Antoine Bordes, Sumit Chopra, and Jason Weston. 2016.
\newblock The goldilocks principle: Reading children's books with explicit
  memory representations.
\newblock In \emph{ICLR}.

\bibitem[{Kadlec et~al.(2016)Kadlec, Schmid, Bajgar, and
  Kleindienst}]{kadlec2016text}
Rudolf Kadlec, Martin Schmid, Ondrej Bajgar, and Jan Kleindienst. 2016.
\newblock Text understanding with the attention sum reader network.
\newblock \emph{arXiv preprint arXiv:1603.01547}.

\bibitem[{Kingma and Ba(2014)}]{kingma2014adam}
Diederik Kingma and Jimmy Ba. 2014.
\newblock Adam: A method for stochastic optimization.
\newblock \emph{arXiv preprint arXiv:1412.6980}.

\bibitem[{Liu et~al.(2017)Liu, Huang, Huang, and Zhang}]{Liu2017}
Zhuang Liu, Degen Huang, Kaiyu Huang, and Jing Zhang. 2017.
\newblock \emph{DIM Reader: Dual Interaction Model for Machine Comprehension}.
  Springer International Publishing, Cham.

\bibitem[{Onishi et~al.(2016)Onishi, Wang, Bansal, Gimpel, and
  McAllester}]{onishi2016did}
Takeshi Onishi, Hai Wang, Mohit Bansal, Kevin Gimpel, and David McAllester.
  2016.
\newblock Who did what: A large-scale person-centered cloze dataset.
\newblock \emph{arXiv preprint arXiv:1608.05457}.

\bibitem[{Paperno et~al.(2016)Paperno, Kruszewski, Lazaridou, Pham, Bernardi,
  Pezzelle, Baroni, Boleda, and Fern{\'{a}}ndez}]{paperno16the}
Denis Paperno, Germ{\'{a}}n Kruszewski, Angeliki Lazaridou, Quan~Ngoc Pham,
  Raffaella Bernardi, Sandro Pezzelle, Marco Baroni, Gemma Boleda, and Raquel
  Fern{\'{a}}ndez. 2016.
\newblock The {LAMBADA} dataset: Word prediction requiring a broad discourse
  context.
\newblock In \emph{{ACL}}.

\bibitem[{Pennington et~al.(2014)Pennington, Socher, and
  Manning}]{pennington2014glove}
Jeffrey Pennington, Richard Socher, and Christopher Manning. 2014.
\newblock Glove: Global vectors for word representation.
\newblock In \emph{Proceedings of the 2014 conference on empirical methods in
  natural language processing (EMNLP)}, pages 1532--1543.

\bibitem[{Seo et~al.(2016)Seo, Min, Farhadi, and Hajishirzi}]{seo2016query}
Minjoon Seo, Sewon Min, Ali Farhadi, and Hannaneh Hajishirzi. 2016.
\newblock Query-reduction networks for question answering.
\newblock \emph{arXiv preprint arXiv:1606.04582}.

\bibitem[{Srivastava et~al.(2014)Srivastava, Hinton, Krizhevsky, Sutskever, and
  Salakhutdinov}]{srivastava2014dropout}
Nitish Srivastava, Geoffrey~E Hinton, Alex Krizhevsky, Ilya Sutskever, and
  Ruslan Salakhutdinov. 2014.
\newblock Dropout: a simple way to prevent neural networks from overfitting.
\newblock \emph{Journal of machine learning research}, 15(1):1929--1958.

\bibitem[{Taylor(1953)}]{taylor1953cloze}
Wilson~L Taylor. 1953.
\newblock “cloze procedure”: a new tool for measuring readability.
\newblock \emph{Journalism Bulletin}, 30(4):415--433.

\bibitem[{Trischler et~al.(2016)Trischler, Ye, Yuan, and
  Suleman}]{trischler2016natural}
Adam Trischler, Zheng Ye, Xingdi Yuan, and Kaheer Suleman. 2016.
\newblock Natural language comprehension with the epireader.
\newblock \emph{arXiv preprint arXiv:1606.02270}.

\bibitem[{Wang et~al.(2015)Wang, Bansal, Gimpel, and
  McAllester}]{wang2015machine}
Hai Wang, Mohit Bansal, Kevin Gimpel, and David McAllester. 2015.
\newblock Machine comprehension with syntax, frames, and semantics.
\newblock In \emph{Proceedings of the 53rd Annual Meeting of the Association
  for Computational Linguistics and the 7th International Joint Conference on
  Natural Language Processing (Volume 2: Short Papers)}, volume~2, pages
  700--706.

\bibitem[{Wang et~al.(2017)Wang, Onishi, Gimpel, and
  McAllester}]{wang2017emergent}
Hai Wang, Takeshi Onishi, Kevin Gimpel, and David McAllester. 2017.
\newblock Emergent predication structure in hidden state vectors of neural
  readers.
\newblock In \emph{Proceedings of the 2nd Workshop on Representation Learning
  for NLP}, pages 26--36.

\bibitem[{Wiseman et~al.(2016)Wiseman, Rush, and Shieber}]{wiseman16learning}
Sam Wiseman, Alexander~M. Rush, and Stuart~M. Shieber. 2016.
\newblock Learning global features for coreference resolution.
\newblock In \emph{{NAACL} {HLT}}, pages 994--1004.

\bibitem[{Wiseman et~al.(2015)Wiseman, Rush, Shieber, and
  Weston}]{wiseman15learning}
Sam Wiseman, Alexander~M. Rush, Stuart~M. Shieber, and Jason Weston. 2015.
\newblock Learning anaphoricity and antecedent ranking features for coreference
  resolution.
\newblock In \emph{Proceedings of the 53rd Annual Meeting of the Association
  for Computational Linguistics ({ACL})}, pages 1416--1426.

\end{thebibliography}
\bibliographystyle{acl_natbib_nourl}

\end{document}